\documentclass[journal]{IEEEtran}

\usepackage[table]{xcolor}
\usepackage{graphicx} 
\usepackage[utf8]{inputenc}
\usepackage{amsmath}
\usepackage{amsfonts}
\usepackage{amsthm}
\usepackage{multicol}
\usepackage{multirow}
\usepackage{array}
\usepackage{hyperref}
\usepackage[margin=1.0in]{geometry}
\usepackage[noend]{algpseudocode}
\usepackage{algorithm}
\usepackage{wrapfig}
\usepackage{setspace}
\usepackage[symbol]{footmisc}
\usepackage{tikz}

\newtheorem{theorem}{Theorem}
\newtheorem{definition}[theorem]{Definition}
\newtheorem{assumption}[theorem]{Assumption}

\renewcommand{\thetable}{\arabic{table}}

\newcolumntype{C}[1]{>{\centering\arraybackslash}p{#1}}

\title{Rapidly-Iterating Grid-Based Near-Optimal Kinodynamic Motion Planning}
\author{Michael Moncton and Eric Frew}
\date{April 2023}

\begin{document}

\maketitle

\begin{abstract}
This paper develops the Rapidly-iterating kino\-Dynamic Grid (RDG) algorithm, an asymptotically near-optimal kinodynamic motion planning algorithm that produces high quality solutions through rapid iteration.
The algorithm leverages a state space grid decomposition to perform node selection, dynamics propagation, and graph revision in constant time complexity with respect to the number of nodes in the trajectory tree.
Through a covering ball sequence induction proof, the algorithm is shown to be asymptotically near-optimal and probabilistically complete. 
Different subsystems of the algorithm are evaluated against common nearest-neighbor search-based methods at generating exploration bias.
The RDG algorithm is evaluated through simulated trials in complex, kinodynamic motion planning problem environments up to 10 DOF relative to similar sparse, kinodynamic planning algorithms with optimality guarantees, SST and DIRT.
The RDG algorithm outperforms both SST and DIRT in mean final solution quality by up to 104\% and 40\% respectively.  
Additionally, RDG maintained a 100\% success rate, even on a 10-DOF test case where both SST and DIRT did not.






%
\end{abstract}

\begin{IEEEkeywords} 
Motion planning, kinodynamic planning, asymptotic near-optimality, exploration bias.
\end{IEEEkeywords}

\section{Introduction}
Sampling-based motion planning (SBMP) algorithms are a common approach for motion planning problems, especially for high dimensional problems \cite{Kavraki1996ProbabilisticSpaces}. 
By randomly sampling feasible trajectories, many SBMP algorithms construct trees of feasible trajectories that explore the admissible state space \cite{LaValle2001RandomizedPlanning}. 

Many such SBMP algorithms plan using a three phase structure: sampling, propagation, and graph revision \cite{LaValle2001RandomizedPlanning}.
First a node in the tree is selected.
That node is then propagated and the graph is revised based on the resulting trajectory segment.
While many algorithms share this basic structure, each of the components can be implemented with a wide variety of methods \cite{Kalisiak2006RRT-blossom:Behavior,Hassidof2025Train-OnceTrees}. 

Some SBMP algorithms are able to converge upon optimal solution trajectories via a property called asymptotic optimality (AO) \cite{Karaman2011Sampling-basedPlanning}.
Specifically, many AO SBMP algorithms utilize a steering function that can optimally connect two individual states \cite{Gammell2020BatchSearch}.
This process involves the solution of a two-point boundary-value problem (BVP) \cite{Osborne1969OnProblems}.
While the solution of the BVP allows optimal paths to be created in a piecewise manner, BVPs are notoriously difficult to solve for some problems.

Solving a BVP, especially with systems that involve non-linear dynamic constraints or have black-box dynamics, can be intractable, especially online. 
Some such problems are referred to as kinodynamic motion planning problems: problems with both kinematic and dynamic constraints \cite{Donald1993KinodynamicREIF}.
One approach to the kinodynamic planning problem is to forward propagate system dynamics via randomly sampled control inputs. 
This approach avoids the complex BVP and is compatible with black-box dynamics \cite{LaValle2001RandomizedPlanning}.
Approaching optimality therefore becomes significantly more challenging in the kinodynamic case.
While AO has been approached by repeated querying of feasible planners, achieving AO with rapid convergence remains a difficult problem \cite{Hauser2016AsymptoticallySpace}.

However, a class of kinodynamic SBMP algorithms exists that relaxes AO guarantees to those of asymptotic near-optimality (ANO) in exchange for performance improvements \cite{Li2016AsymptoticallyPlanning}.
Such algorithms have probabilistic guarantees on sampling trajectories which have a bounded cost difference from an optimal trajectory.

\begin{figure*}[t]
\centering
\includegraphics[width=.9\linewidth]{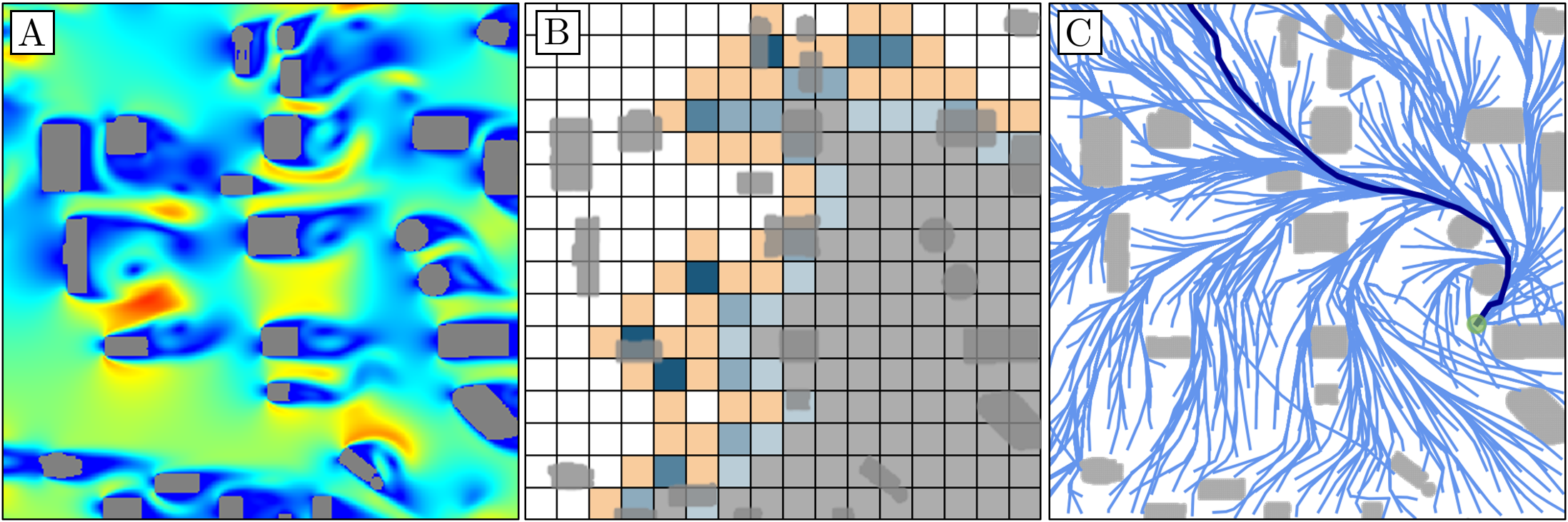}
\caption{A. The RDG planner is suited for complex kinodynamic or black-box dynamics problems. B. Grid-based methods allow for constant time generation of exploration bias. C. Local continuous trajectory refinement within the grid creates asymptotically near-optimal solutions.}
\label{fig:intro}
\end{figure*}

Without solving for optimal trajectory segments themselves, planners must rely on randomly sampling high quality trajectory segments.
For all but the simplest planning problems, the probability of sampling a feasible solution trajectory, let alone a high quality solution trajectory, is very low.
For ANO planners that forward propagate dynamics, the number of trajectory samples is therefore critically important \cite{Shome2021AsymptoticallyEdges}.
Without a sufficiently large number of samples, the probability that a high-quality solution is found is small. 
Then for online operation, the rate at which the planner can sample trajectories becomes a critically important quantity. 

To improve sampling rate and convergence speeds, the Stable Sparse RRT (SST) algorithm applies the ANO framework to a sparse trajectory tree \cite{Li2016AsymptoticallyPlanning}. 
SST maintains this tree via the use of a pruning function, which only accepts nodes that have a low cost relative to other nodes in a nearby ``witness," a state-space neighborhood where cost-to-arrive is approximated by the cost of a single node. 
To utilize witnesses in the selection and graph revision stages of the algorithm, nearest-neighbor search (NNS) and fixed-radius near-neighbors search (FRNNS) operations are utilized. 
The computational cost of these operations scales at best logarithmically and linearly respectively with the number of active nodes in the tree and has been shown to be competitive with collision checking in computational cost for certain problems \cite{Kleinbort2020CollisionPlanning}.
The Dominance-Informed Region Tree (DIRT) algorithm can utilize a similar sparse structure (DIRT-prune), while improving upon the performance of SST by replacing witnesses with dynamically sized dominance-informed regions, utilizing a multi-sample blossom propagator, and taking advantage of a guidance heuristic \cite{Littlefield2018EfficientRegions}.
However, similar to SST, the utilization of DIRs still requires NNS and FRNNS operations. 

As the SST and DIRT algorithms grow their respective trees, the rate of trajectory sampling of both algorithms slows significantly. 
The Rapidly-iterating kinoDynamic Grid (RDG) algorithm replaces both witnesses and DIRs with a uniform grid spanning the state space, which allows for selection and revision subroutines to be performed in $O(1)$ complexity with respect to the number of active nodes in the tree. 

Recent work has begun to remove the NNS bottleneck from the geometric AO planning space through parallelism \cite{Wilson2025Nearest-NeighbourlessTrees}.
Grid-based planning itself has also been utilized in the motion planning literature. 
The KPIECE algorithm uses grid decompositions of the state space to track exploration \cite{Sucan2009KinodynamicExploration}. 
Grid-based selection and graph revision methods have been used in the context of kinodynamic planning problems for parallelization purposes \cite{Perrault2025Kino-PAX:Planner}.
While the resulting Kino-PAX algorithm is able to take advantage of highly parallelized GPU computing to make gains in planning speed, it does not solve the time-complexity problem, but rather utilizes parallelization to reduce its significance.
While both KPIECE and Kino-PAX provide a baseline of existing methodology on grid-based kinodynamic planning, KPIECE does not provide optimality guarantees and Kino-PAX does not explicitly address the time-complexity problem.

In some recent motion planning literature, learning assisted planning has been used to accelerate solution finding. 
Diffusion policies have been used in place of random trajectory sampling in the propagation component of a planner \cite{Hassidof2025Train-OnceTrees}.
While doing so provides strong planner speed increases for unstable systems due to more efficient trajectory sampling, expert trajectory data is required and does not necessarily generalize to other systems without retraining. 

It was also shown in \cite{Hassidof2025Train-OnceTrees} that as long as a propagation method is fully supported over the control space it is compatible with an array of selection and graph revision methods, such as those of RRT and SST.
Similarly, the ANO proof in this work requires a fully-supported propagation function.

Additionally, end-to-end learned planning has been used to return trajectories without the need for the aforementioned three part structure.
By learning the planning problem itself, entire solution trajectories can be returned in the time required to query a neural network \cite{Kicki2024FastNetworks}.
However, trajectories returned by such methods are not guaranteed to be feasible. 

In this paper, we present the RDG algorithm, which utilizes grid-based selection and graph revision methods to ANO kinodynamic planning.
By replacing the witnesses and DIRs of SST and DIRT respectively with a uniform grid, RDG achieves O(1) per-iteration time complexity with respect to the number of active nodes in the tree. 
While the propagator used here forward propagates uniform samples from the control space, any fully-supported state propagator satisfies the theoretical assumptions required for ANO.
RDG does not require expert data, and relies only on forward propagation of dynamics, making it applicable to problems where learned alternatives are not available. 
The RDG algorithm is validated against the state-of-the art in its specific class of algorithm, SST and DIRT, across one geometric and four kinodynamic test cases.

\section{Problem and Definitions}
\subsection{Definitions}
This work assumes a bounded state space $X$.
Let the admissible subset of the state space $X_A$ be
\begin{equation}
X_A = X\backslash X_{I}
\end{equation}
where $X_{I} \subset X$ is the inadmissible set, e.g. due to obstacles, self-collision, performance envelope, etc. Let $\mathcal{B}_\delta(x)$ be the closed ball in the state space with radius $\delta$ centered at $x \in X$. 
Consider a system that evolves according to 
\begin{equation} \label{eq:dynamics}
\dot{x} = f(x,u)
\end{equation}
where $x \in X$, and $u \in U$, the control space.

\begin{definition}[Trajectory, Definition 1 in \cite{Li2016AsymptoticallyPlanning}]
A trajectory $\pi$ is a mapping $\pi(t):[0,t_\pi]\rightarrow X_A$ generated by integrating Equation \ref{eq:dynamics} with a control sequence $u(t)$.
\end{definition}

Trajectories are often formed as a concatenation of trajectory segments.
Each trajectory segment $\pi_k$ is created by starting at the initial state of the segment and integrating the system dynamics in Equation \ref{eq:dynamics} using a control sequence $u(t)$.
The cost of a trajectory, $\texttt{cost}(\pi)$, is assumed to be Lipschitz continuous, additive, monotonic, and non-degenerate, as in Assumption 11 in \cite{Li2016AsymptoticallyPlanning}. 
An optimal trajectory therefore connects an initial and final state through a control function that minimizes cost. 
A trajectory segment is considered optimal if it connects initial and final states with minimum cost.
Consider a motion planning problem with an optimal solution cost $\texttt{cost}(\pi^*) = c^*$.
Given an initial state $x_0$ and a goal region $X_{goal} \subset X$, the optimal kinodynamic motion planning problem seeks to find a trajectory $\pi^*(t) \subset X_A, \forall t \in [0,t_{\pi^*}]$ where $\pi^*(0) = x_0$ and $\pi^*(t_{\pi^*})\in X_{goal}$ that minimizes cost. 

For a motion planning problem to be solvable by a sampling-based algorithm, $\pi^*$ must satisfy certain clearance properties that ensure that the probability of sampling a solution trajectory is greater than zero. 
\begin{definition}[Obstacle Clearance]
The obstacle clearance $\delta_{obs}$ of a trajectory $\pi$ is the minimum distance between any point on the trajectory and any inadmissible state $x \notin X_A$:
\begin{equation}
\delta_{obs}(\pi) = \min\big(|x-x_\pi|\big), \forall x_\pi \in \pi, \forall x\notin X_A
\end{equation}
\end{definition}

While it is referred to as obstacle clearance, this term is defined slightly more generally to include clearance from any inadmissible state. 
Such states may be inadmissible but not in obstacles due to state-space boundaries such as self-collision, imposed performance limits, etc.

\begin{definition}[Dynamic Clearance, Lemma 6 in \cite{Li2016AsymptoticallyPlanning}]
A path $\pi(t)$ has dynamic clearance $\delta$ if, for any two points $x_i = \pi(t_i), \; x_{i+1} =\pi(t_{i+1}), \; t_{i+1} > t_i$,  $\forall x_i' \in \mathcal{B}_\delta(x_i)$ and $\forall x_{i+1}' \in \mathcal{B}_\delta(x_{i+1})$ there exists a trajectory $\pi'$ such that $\pi'(0) = x_{i}'$ and $\pi'(t_{\pi'}) = x_{i+1}'$.
\label{def:dynclear}
\end{definition}

\begin{definition}[Near-Optimal Kinodynamic Motion Planning Problem]
Given an initial state $x_0$ and a goal region $X_{goal} \subseteq X_A$ the near-optimal kinodynamic motion planning problem seeks to find a trajectory $\pi$ with cost $c \leq h(c^*,\delta)$ where $c^*$ is the cost of the optimal solution trajectory $\pi^*$:
\begin{equation}
\pi^* = \underset{\pi \in \Pi}{argmin}(\texttt{cost}(\pi))
\end{equation}
where $\Pi$ is the set of all admissible solution paths.
\end{definition}

The near-optimal motion planning problem seeks to find a trajectory that is of bounded cost difference from the optimal solution cost.
This cost bound, $h(c^*,\delta)$, is a function of the optimal solution path cost and the dynamic clearance of the optimal solution trajectory, $\delta$. 

For a trajectory $\pi$ to have dynamic clearance $\delta$, every state $x_i'$ in the $\delta$ ball around $x_i$ must be able to reach every state $x_{i+1}'$ in the $\delta$ ball around $x_{i+1}$.

\begin{definition}[$\delta$-Robust Trajectories, Definition 7 in \cite{Li2016AsymptoticallyPlanning}]
A trajectory $\pi$ is considered $\delta$-robust if both its obstacle clearance and dynamic clearance are greater than $\delta$.
\end{definition}

The $\delta$-robustness of $\pi^*$ is a necessary property for many sampling-based motion planning algorithms to achieve probabilistic completeness.
It is useful to note that obstacle clearance cannot be smaller than dynamic clearance as the existence of trajectories in $\mathcal{B}_\delta$ implies that $\mathcal{B}_\delta$ is obstacle free.

\begin{definition}[$\delta$-Similar Trajectories, Definition 3 in \cite{Li2016AsymptoticallyPlanning}]
Trajectories $\pi$ and $\pi'$ are $\delta$-similar if, for a continuous non-decreasing scaling function $\sigma : [0,t_\pi] \rightarrow [0,t_{\pi'}]$, $\pi'(\sigma(t)) \in \mathbb{B}_\delta(\pi(t))$.
\end{definition}

Paths that are $\delta$-similar will have a maximum distance of separation $\delta$ from any point on one path to the other.
For a trajectory with dynamic clearance of $\delta$, any two balls constructed on the trajectory can be connected to each other. 
Therefore, trajectories that are $\delta$-robust are guaranteed to have feasible $\delta$-similar trajectories.
For this work, trajectories that are $\delta$-similar to $\pi^*$ are of primary concern. 

\begin{assumption}[Cost Function (Assumption 11 in \cite{Li2016AsymptoticallyPlanning})]
The cost of a trajectory $\pi$, $\texttt{cost}(\pi)$, is assumed to be Lipschitz continuous. Specifically, $\exists \; K_c > 0$:
\begin{equation}
|\texttt{cost}(\pi_0) - \texttt{cost}(\pi_1)| \leq K_c*\sup_{\forall t}\big\{||\pi_0(t) - \pi_1(t)||\big\}
\end{equation}
for all $\pi_0$ and $\pi_1$ with the same initial state. Consider two trajectories $\pi_0$ and $\pi_1$ such that their concatenation is $\pi_0|\pi_1$. The cost functions satisfies:
\begin{enumerate}
    \item $\texttt{cost}(\pi_0|\pi_1) = \texttt{cost}(\pi_0) + \texttt{cost}(\pi_1)$,
    \item $\texttt{cost}(\pi_0) < \texttt{cost}(\pi_0|\pi_1)$, and
    \item $\forall\; t_2 > t_1 \geq 0, \exists \; M_c > 0$ so that $t_2 - t_1 \leq M_c \cdot|\texttt{cost}(\pi(t_2)) - \texttt{cost}(\pi(t_1))|$
\end{enumerate}
which represent the properties of additivity, monotonicity, and non-degeneracy respectively.
\label{assump:cost}
\end{assumption}

The properties of additivity, monotonicity, and non-degeneracy are important properties of trajectories that are commonly considered in optimal motion planning problems.

\begin{definition}[Probabilistic $\delta$-Robust Completeness]
If an algorithm is probabilistically $\delta$-robustly complete, the probability that the algorithm finds a solution as time approaches infinity is 1 provided that a $\delta$-robust solution exists.
\end{definition}

Probabilistic completeness is an important property of sampling-based motion planning algorithms. 
Probabilistic $\delta$-robust completeness extends this concept to problems concerning $\delta$-robust solutions where non-$\delta$-robust solutions may be impossible to find.

\begin{definition}[Asymptotic $\delta$-Robust Near-Optimality (Definition 12 in \cite{Li2016AsymptoticallyPlanning})]
Let $c^*$ denote the minimum cost over all solution trajectories for a $\delta$-robust feasible motion planning problem. Let $Y^{ALG}_n$ denote a random variable that represents the minimum cost value among all trajectories returned by algorithm ALG at iteration $n$ for the same problem. ALG is asymptotically $\delta$-robustly near-optimal if:
\begin{equation}
P\bigg(\Big\{\underset{n\rightarrow \infty}{\lim \sup }\; Y^{ALG}_n \leq h(c^*,\delta)\Big\}\bigg) = 1
\end{equation}

Furthermore, any admissible trajectory that connects $x_0$ and $X_{goal}$ with cost less than $h(c^*,\delta)$ is a near-optimal trajectory $\pi_{NO}$.
\label{def:ANO}
\end{definition}

Asymptotic $\delta$-robust near-optimality is a property of an algorithm concerning the cost of solutions that the algorithm will find.
This property states that the algorithm will find a solution with a cost bounded by $h(c^*,\delta)$ asymptotically almost surely.
Notably, such trajectories do not necessarily have to be $\delta$-similar to $\pi^*$ as long as their cost is sufficiently low. 

\section{The Rapidly-iterating kinoDynamic Grid Algorithm}

This section describes the Rapidly-iterating kinoDynamic Grid (RDG) algorithm. To initialize the RDG algorithm, several inputs are required:
the starting state $x_0$, a goal region $X_{goal}$, a uniform, axis-aligned, hyper-rectangular cell grid $\mathbb{C}$, the dynamics function necessary for propagation, and the cost function $\texttt{cost}(\pi)$.

\subsection{Planner Structures}
The RDG algorithm relies on two graphs: a trajectory tree $G(\mathbb{V},\mathbb{E})$, and the cell grid $\mathbb{C}$.
The trajectory tree is rooted in $v_0$, the root node located at $x_0$ with $\texttt{cost}(v_0) = 0$, and consists of connected kinodynamically feasible trajectory segments.
A node in the tree will be referred to as $v_i$ and is located at state $x_i$.
This distinction exists to differentiate between nodes that may reach the same location in the state space via different trajectories.
The cell grid spans the state space $X$ and defines cells $c \in \mathbb{C}$.

\begin{figure}
\centering
\includegraphics[width=.7\linewidth]{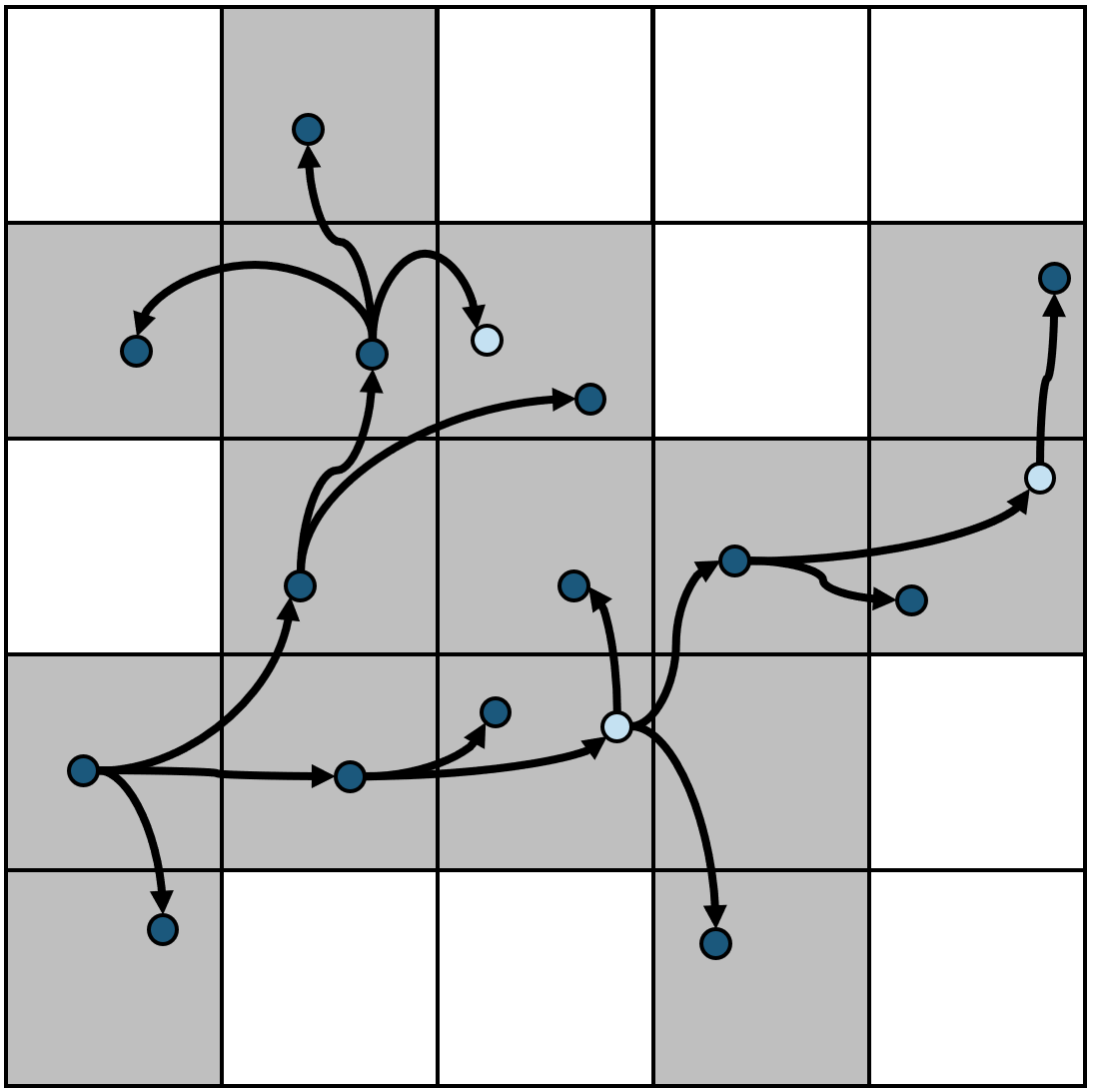}
\caption{Planner structures include trajectories (black lines), active nodes (dark circles), inactive nodes (light circles), and explored cells (gray)}
\label{fig:plannerstructures}
\end{figure}

Adjacent cells are defined as orthogonally adjacent, meaning that in a state space in $\mathbb{R}^n$, each cell has at most $2n$ neighbors.
Cells are considered overlapping on the borders and a node that lies exactly on said border is evaluated in both cells independently. 
If a node from the trajectory tree exists inside a cell $c$ in the cell grid, $c$ is considered \textit{explored} and is otherwise \textit{unexplored}.
The node in $c$ with the lowest cost is referred to as the representative node of $c$, $c.rep$.
The representative node of $c$ is referred to as an \textit{active node} and is able to be expanded via propagation.
Each explored cell has exactly one active node at any given time.
Other nodes inside $c$ with worse cost are \textit{inactive nodes} and will not be propagated further.
Inactive nodes, however, are not deleted. 
Even if a node does not have the best cost inside its local cell, it is possible that it has active child nodes. 
An example trajectory tree and cell grid can be seen in Figure \ref{fig:plannerstructures}.

\subsection{Algorithm Overview}
The RDG algorithm (Algorithm \ref{alg:cap}) follows the three-part structure shared by many SBMP algorithms.
In the first step, the \Call{Node\_Selection}{} sub-algorithm selects an active node $v_{prop}$ from the trajectory tree for expansion.
The \Call{Propagation}{} sub-algorithm then generates a random trajectory segment originating from $v_{prop}$ and terminating in a new node $v_{new}$ using a random control input applied for a bounded random time $0 < t_{prop} < t_{prop,max}$.
If the resulting node $v_{new}$ and its associated trajectory are collision free, the \Call{Graph\_Revision}{} sub-algorithm evaluates $v_{new}$ by comparing it to the active node in the local cell of $x_{new}$ if one exists.
The cell grid is also revised in this step, potentially marking unexplored cells as explored or assigning $v_{new}$ as a new active representative node. 

On initialization, the starting state $v_0$ is assigned as the active representative of its local cell and that cell is marked as explored. 
Data structures for tracking neighboring unexplored cell counts are also initialized.

\begin{algorithm}
\caption{\texttt{Rapidly-iterating kinoDynamic Grid}}\label{alg:cap}
\begin{algorithmic}[1]
\State $G(\mathbb{V},\mathbb{E}) \gets \Call{Initialize\_Tree}{x_0}$
\For {$N\;iterations$}{}

\State $v_{prop} \gets$ \Call{Node\_Selection()}{}
\State $v_{new} \gets$ \Call{Dynamics\_Propagation}{$x_{prop}$}
\If{\Call{Collision\_Free}{$v_{prop}\rightarrow v_{new}$}}
\State \Call{Graph\_Revision}{$v_{new}$}
\EndIf
\EndFor
\end{algorithmic}
\end{algorithm}

\subsection{Node Selection}

\begin{algorithm}
\caption{$\texttt{Node Selection}$}
\begin{algorithmic}[1]
\State $c_{rand} \gets$ \Call{Random\_Cell()}{}
\If{$c_{rand}$ is explored}
\Return{$c_{rand}.rep$}
\EndIf
\State \Return{\Call{Selection\_Function}{}}
\end{algorithmic}
\label{alg:nodeselection}
\end{algorithm}

The $\Call{Node\_Selection}{}$ function in Algorithm \ref{alg:nodeselection} selects a node for propagation by the planner. 
A random cell $c_{rand}$ is selected from the grid. 
If there exists a node in $c_{rand}$ already, then the active node in $c_{rand}$ is returned. 
If $c_{rand}$ is an unexplored cell, a \Call{Selection\_Function}{} is used to select a node with the purpose of exploring the environment.
As the number of explored cells increases, the probability of calling the selection function decreases, and reaches zero if all cells are explored.
This shift allows node selection to be less and less exploration-biased as the environment becomes more explored.

\begin{algorithm}
\caption{\texttt{Random Frontier}}
\begin{algorithmic}[1]
    \State $C_{exp} = \{c \in \mathbb{C}\; |\; c.rep \neq null \}$
    \State $C_f = \{c \in C_{exp}\; |\; \exists \; c.neighbor.rep = null \}$
    \State $c_{select} \gets C_f.get\_random()$
    \State \Return $c_{select}.rep$
\end{algorithmic}
\label{alg:rf}
\end{algorithm}

One potential selection function picks nodes that are near unexplored cells.
An explored cell that is adjacent to an unexplored cell is referred to as a frontier cell and the adjacent unexplored cell a boundary cell. 
Selecting the active representative of a frontier cell is more likely to result in a propagation that explores new cells than selecting an interior cell that is surrounded by already explored cells.
This method is the $\Call{Random\_Frontier}{}$ selection function in Algorithm \ref{alg:rf}.

\begin{figure}
\centering
\includegraphics[width=.75\linewidth]{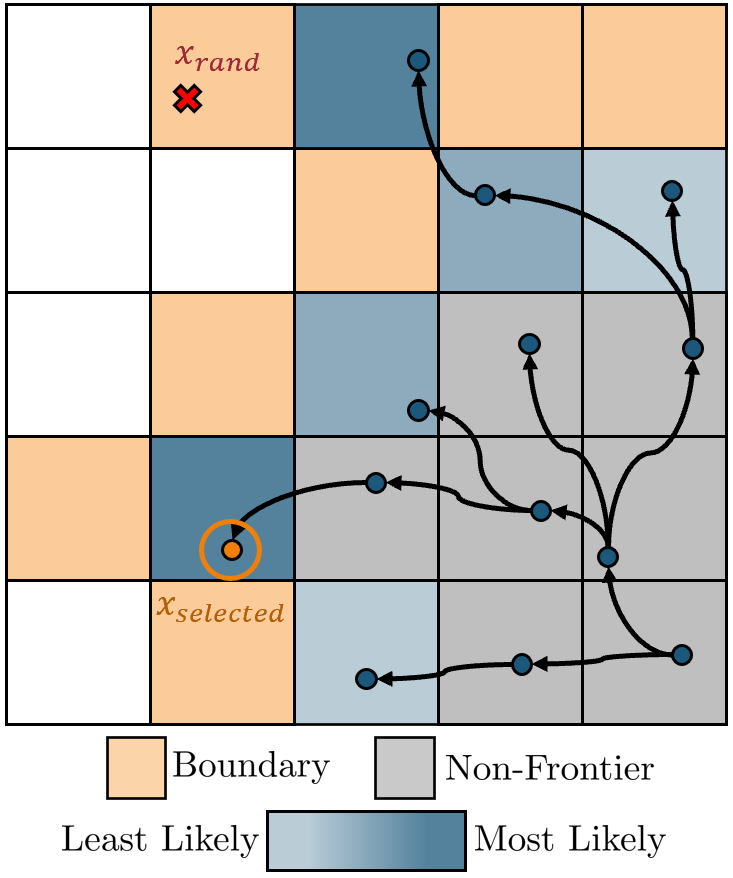}
\caption{The weighted random frontier node selection method}
\label{fig:stateselection}
\end{figure}

\begin{algorithm}
\caption{\texttt{Adjacency-Weighted Random Frontier}}
\begin{algorithmic}[1]
    \State $C_{exp} = \{c \in \mathbb{C}\; |\; c.rep \neq null \}$
    \State $C_f = \{c \in C_{exp}\; |\; \exists \: c.neighbor.rep = null \}$ 
    \State \parbox[t]{.8\linewidth}{%
    $c_{select} \gets C_f.get\_random()$ such that \\
    $P(c)\; \propto\; c.n\_unexplored\_neighbors$}
    \State \Return $c_{select}.rep$
\end{algorithmic}
\label{alg:wrf}
\end{algorithm}

This process can be further improved by weighting the selection of frontier cells by the number of neighboring boundary cells. 
Selecting a frontier cell with many neighboring boundary cells is more likely to result in exploration of a new cell than selecting a cell with only a single neighboring boundary cell.
The RDG algorithm uses the weighted random frontier (WRF) method in Algorithm \ref{alg:wrf} as its selection function. Further analysis of the efficacy of both Algorithm \ref{alg:rf} and Algorithm \ref{alg:wrf} can be found in Section \ref{sec:Ebias}.

\subsection{Graph Revision}
The purpose of graph revision in the planning algorithm is to maintain the sparsity of the tree.
If two trajectory segments terminate near each other but have very different path costs, the graph revision algorithm seeks to only maintain the node with lower cost. 
The cell grid is used to accomplish this without compromising the time complexity of the algorithm.
The graph revision algorithm for RDG can be found in Algorithm \ref{alg:graphrevision}.

\begin{algorithm}
\caption{\texttt{Graph Revision}$(v_{new})$}
\begin{algorithmic}[1]
\State $c_{local} \gets$ \Call{Local\_Cell}{$x_{new}$}
\If{$c_{local}.rep = null$ }
    \State $c_{local}.rep \gets v_{new}$
    \State add $v_{new}$ to graph
\ElsIf{$v_{new}.cost < c_{local}.rep.cost$}
    \State $c_{local}.rep \gets v_{new}$ 
    \State add $v_{new}$ to graph
\EndIf
\end{algorithmic}
\label{alg:graphrevision}
\end{algorithm}

After a newly propagated node $v_{new}$ is generated and its associated source trajectory segment is determined to be collision free, its local cell $c_{local}$ is found. 
If $c_{local}$ is unexplored, $v_{new}$ is set as the representative of $c_{local}$. 
If $c_{local}$ was already explored, the cost of $v_{new}$ is evaluated against the cost of the representative of $c_{local}$. 
The representative node of $c_{local}$ is set to the node with the lower cost.
Inactive nodes remain in the graph and corresponding data structure, but are not propagated throughout the rest of the algorithm. 
Keeping inactive nodes saves computation time that is associated with deletion from the data structure.

\subsection{Algorithm Time Complexity}
One of the main features of the selection and revision methods described in this section is that both are performed in constant time complexity with respect to the number of nodes in the graph.
The time complexity of each individual iteration of the RDG algorithm can be evaluated by examining each of the three sub-algorithms that compose it.
Node selection is composed of both the cell occupancy check in lines 1-3 of Algorithm \ref{alg:nodeselection} and the WRF method in Algorithm \ref{alg:wrf}. 
The propagation function does not perform operations over the node set and is therefore in $O(1)$.
Similarly, collision checking is in $O(1)$.
Graph revision, in Algorithm \ref{alg:graphrevision}, requires a $\texttt{Local\_Cell}$ function call, which can easily be performed in constant time for an axis-aligned, hyperrectangular cell grid.
Additionally, the data structure responsible for WRF sampling is updated after new cells are explored, which scales linearly with problem dimensionality, but not with tree size.
The WRF method samples from a weighted random distribution, as well as modifies said distribution in $O(1)$ time.
The data structure that allows such operations maintains a list of frontier cells as well as their weights, taking advantage of the fact that each cell has a discrete, bounded, and constant number of neighbors known at planner initialization.

\section{Asymptotic Near-Optimality}
This section establishes the asymptotic near-optimality of the RDG algorithm.
By Definition \ref{def:ANO}, a near-optimal solution has a cost that is bounded by a function of the dynamic clearance $\delta$ and optimal cost $c^*$. 
A trajectory that is $\delta$-similar to $\pi^*$ is near-optimal \cite{Li2016AsymptoticallyPlanning}.
However, $\delta$-similarity is not a necessary condition to guarantee that a trajectory is near optimal.
By generating nodes that are better than at least one $\delta$-similar path in every cell along the optimal path, the algorithm will generate a near-optimal solution asymptotically almost surely.
 
\subsection{Preliminaries}

This section provides several definitions and assumptions needed before presenting the proof.

\begin{definition}[Covering Cell Sequence]
For a trajectory $\pi$, the covering cell sequence $C(\pi)$ is the ordered set of $n$ cells that $\pi$ passes through: $C(\pi) = \{c_0, c_1,...,c_{n-1}\}$ such that $\forall x_i \in \pi$, $x_i \in C(\pi)$ and $\forall c_i \in C(\pi), \exists\; x_i \in c_i$.
\end{definition}
In the case that the path reenters a cell that it previously exited, the cell exists multiple times in the covering cell sequence.
The following analysis holds regardless.

\begin{definition}[Covering Ball Sequence, Definition 14 in \cite{Li2016AsymptoticallyPlanning}]
For a trajectory $\pi(t)$ and a cost difference value $\Delta_{cost} > 0$, the covering ball sequence $\mathbb{B}_\delta(\pi)$ is the set of $M+1$ hyper-balls $\{\mathcal{B}_\delta(x_0), \mathcal{B}_\delta(x_1),..., \mathcal{B}_\delta(x_M) \}$ where $x_i \in \pi$ and $\texttt{cost}(x_{i+1})-\texttt{cost}(x_i) = \Delta_{cost}$ for $i = 0,1,...,M-1$.
\end{definition}

The size of the cell decomposition is also critical to the properties of the algorithm. 

\begin{assumption}
For a $\delta$-robust planning problem, the diagonal length of each cell is less than $\delta$. 
\label{thm:cellsize}
\end{assumption}

This assumption implies that $\forall x_i^* \in \pi^*$, $\mathcal{B}_\delta(x_i^*)$ is guaranteed to fully contain at least one cell $c_i \in C(\pi^*)$.
Then, there must exist some value of $\Delta_{cost}$ such that every cell in $C(\pi^*)$ is contained in at least one ball, ensuring that $C(\pi^*) \subset \mathbb{B}_\delta(\pi^*$). 
Functionally, this assumption governs the relationship between the dynamic clearance of the optimal solution path and near-optimal solution paths.
While the cell diagonal can be set to be close to $\delta$, setting the cell size smaller will change the resolution of the covering cell sequence and the maximum difference between the optimal solution path and near-optimal paths.
The dynamic clearance of $\pi^*$, however, is the upper bound on the cell size while maintaining the properties of the planning algorithm.

\begin{definition}[Semi-$\delta$-Similar Node]
A node $v$ is a semi-$\delta$-similar node if the trajectory that connects $v_0$ to $v$ is $\delta$-similar to a continuous subset of $\pi^*$: $\{\pi^*(t) | t \in [0,t_s]\},\;t_s \leq t_\pi$.
\label{def:SDSnode}
\end{definition}
To generate a $\delta$-similar trajectory, semi-$\delta$-similar nodes must be repeatedly sampled and propagated into segments that are $\delta$-similar to segments of $\pi^*$. 

\begin{definition}[Near-Optimal Node]
A node $v$ in $G(\mathbb{V},\mathbb{E})$ is a near-optimal node if it is part of a near-optimal trajectory, i.e. $x_v \in \pi_{NO}$.
\end{definition}
While a $\delta$-similar node can be shown to be a near-optimal node, a near-optimal node is not necessarily a $\delta$-similar node.

\subsection{Proof of Asymptotic Near-Optimality}
This section will prove asymptotic near-optimality by showing that the RDG algorithm will probabilistically generate near-optimal nodes in every ball in a covering ball sequence about the optimal solution path. 
If the resulting solution trajectory lies continuously within the covering cell sequence, it also lies within $\mathbb{B}_\delta(\pi^*)$ and is $\delta$-similar to $\pi^*$ and therefore near-optimal.
It will also be shown that near-optimal nodes that are formed from non-$\delta$-similar trajectories can still result in near-optimal solutions.

Let $v_i$ be an active, near-optimal node in cell $c_{i}\in C(\pi^*)$.
Consider one iteration of the algorithm where it attempts to propagate a trajectory from $v_i$ into the next cell in the sequence, $c_{i+1}$.
Two events must occur for this propagation to be created: the node $v_i$ must be selected by Algorithm \ref{alg:nodeselection} and a propagation must be sampled from $v_i$ into $c_{i+1}$.

\begin{theorem}
Given a state space of measure $\mu(X)$ and a cell grid with individual cells of measure $\mu(c_i) \leq \mu(X)$, Algorithm \ref{alg:nodeselection} has a probability of selecting any active node $P_{s} \geq \frac{\mu(c_i)}{\mu(X)} > 0$.
\label{thm:select}
\end{theorem}

Once a node exists within a cell $c_i$, there is guaranteed to always be an active node within that cell for all future planner iterations.
The active representative node of $c_i$ can be replaced by a different node, but the algorithm will always maintain an active representative of $c_i$ once one is created. 
With uniform state sampling in Algorithm \ref{alg:nodeselection} line 1, lines 2 and 3 guarantee that the representative node of any explored cell has a non-zero probability of being selected.
If a random state is sampled within a cell with an active node, that active node will be selected by Algorithm \ref{alg:nodeselection}.
The probability of selecting any active node can therefore be lower-bounded by the probability of sampling a state inside the cell it represents, which is equal to the ratio of the size of the cell to the size of the state space:
\begin{equation}
P_{s,min} = \frac{\mu(c_i)}{\mu(X)}.
\end{equation}
With the addition of the selection function, this probability can only increase. 
The probability of selecting $v_i$ is therefore greater than zero. \qed 

As this property follows from the first three lines of Algorithm 2, Theorem \ref{thm:select} holds with selection functions other than the weighted random frontier selection method in Algorithm \ref{alg:wrf}.

\begin{figure}
\centering
\includegraphics[width=.95\linewidth]{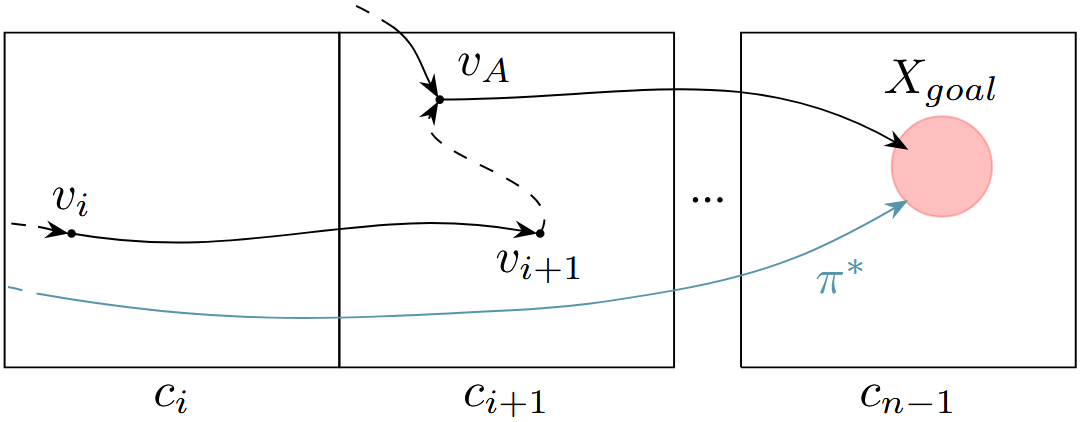}
\caption{The near-optimal node $x_i$ and a new propagation $x_{i+1}$ can be connected to non-$\delta$-similar node $x_A$. While the trajectory ending in $x_A$ is not $\delta$-similar to $\pi^*$, it can still be near-optimal, leading to near-optimal solution paths even if $x_{i+1}$ is potentially rejected.}
\label{fig:proof}
\end{figure}

\begin{theorem}
For the covering cell sequence $C(\pi^*)$, the probability of generating a propagation from a node $v_i \in c_i$ that terminates in cell $c_{i+1}$ is $P_{p}^{(i)} > 0$.
\label{thm:prop}
\end{theorem}

Consider the cell $c_i$ and the next cell in the sequence $c_{i+1}$.
Also consider states $x_i^* \in c_i \cap \pi^*$ and $x_{i+1} \in c_{i+1} \cap \pi^*$.  
It is always possible to construct a covering ball sequence on $\pi^*$ such that covering balls exist centered on points $x_i^*$ and $x_{i+1}^*$. 
Additionally, $c_i \subset \mathcal{B}_\delta(x_i^*)$ and $c_{i+1} \subset \mathcal{B}_\delta(x_{i+1}^*)$ by Assumption \ref{thm:cellsize}.
As both $x_i^*$ and $x_{i+1}^*$ are on $\pi^*$, which is $\delta$-robust, there must exist a trajectory between every point in $\mathcal{B}_\delta(x_i^*)$ and every point in $\mathcal{B}_\delta(x_{i+1}^*)$ by Definition \ref{def:dynclear}.
Therefore, a trajectory must exist between every point in $c_i$ and $c_{i+1}$.
Particularly, for any individual point in $c_i$, there must exist a trajectory between it and every point in $c_{i+1}$.
The set of all such trajectories therefore has dimensionality equal to that of the state space and the probability of sampling one such trajectory is $P_{p}^{(i)} > 0$, as shown in Theorem 17 in \cite{Li2016AsymptoticallyPlanning}. 
\qed

While uniform random sampling is used in the proof of Theorem 17 in \cite{Li2016AsymptoticallyPlanning}, uniform sampling is a sufficient but not necessary condition of the proof.
That proof only requires that the control sampling distribution be fully supported over the admissible control space, not that it be uniform \cite{Hassidof2025Train-OnceTrees}.

From Theorem \ref{thm:select} and Theorem \ref{thm:prop}, the probability of the two respective events, selecting $v_i$ and propagating $v_i$ into $c_{i+1}$ both have non-zero probabilities of occurring.
Therefore, the event of generating a node in $c_{i+1}$ has a non-zero probability of occurring and will happen with a probability of 1 as the number of algorithm iterations goes to infinity.
The next step is ensuring that such a node will be added to the graph.

\begin{theorem}
If a near-optimal node is created in $c_{i+1}$ as the result of a propagation, at the end of the current iteration there will exist a near-optimal active node in cell $c_{i+1}$.
\label{thm:keep}
\end{theorem}
Once the new propagation is generated, Algorithm \ref{alg:graphrevision} determines if the resulting node $v_{i+1}$ is added to the graph or discarded.
For the propagation to be added, one of three possible outcomes must then occur.
\begin{enumerate}
    \item The new node $v_{i+1}$ can be added to the tree, either by replacing an existing node of worse cost in $c_{i+1}$ or by being the first node created in $c_{i+1}$.
    \item The new node can be found to be of worse cost than the active representative of $c_{i+1}$, an existing semi-$\delta$-similar node, in which case it is discarded.
    \item The new node can be found to be of worse cost than the active representative $v_A \in c_{i+1}$ which is not a semi-$\delta$-similar node.
\end{enumerate}
In the first two cases, after the iteration completes, there exists a near-optimal active node in cell $c_{i+1}$, either the newly created node or the one it was rejected in favor of.
The third case presents a challenge, as a node, which was created from a semi-$\delta$-similar propagation from a near-optimal node has been rejected in favor of a non-semi-$\delta$-similar node. 
If $v_A$ is of sufficiently low cost, potentially a cost lower than that of $v_i$, it can prevent the addition of any nodes propagated from $v_i$ terminating in $c_{i+1}$.
In this case, the creation of a node in $c_{i+1}$ from $v_i$ becomes impossible.

However, if $v_{i+1}$ was rejected in favor of $v_A$, it follows that $\texttt{cost}(v_A) < \texttt{cost}(v_{i+1})$. 
Let $\pi_A$ be the trajectory that terminates in $v_{i+1}$ concatenated with a trajectory that connects $v_{i+1}$ to $x_A$.
The connecting trajectory must exist by Definition \ref{def:dynclear}.
One such trajectory can be seen in Figure \ref{fig:proof}.
As the connecting trajectory between $v_i$ and $v_{i+1}$ is $\delta$-similar to a subset of $\pi^*$, it is of bounded cost, making $v_{i+1}$ a near-optimal node. 
By Assumption \ref{assump:cost}, $\texttt{cost}(\pi_A) > \texttt{cost}(v_{i+1})$.
Given that $\texttt{cost}(v_A) < \texttt{cost}(v_{i+1})$, it follows that $\texttt{cost}(v_A) < \texttt{cost}(\pi_A)$.
Therefore, in all three cases a near-optimal node exists in $c_{i+1}$. \qed

This proof shows that, while $\delta$-similar nodes may be rejected by the graph revision subalgorithm, the nodes that they are rejected in favor of are near-optimal themselves, even if they are not $\delta$-similar.
This conclusion is based on the logic in \cite{Moncton2026Achievingdelta-Similarity}, where it was shown that the SST algorithm can maintain asymptotic near-optimality without guarantees about sampling $\delta$-similar paths.

\begin{theorem}
As the number of iterations approaches infinity, the probability of the RDG algorithm generating a near-optimal solution approaches 1.
\label{thm:ano}
\end{theorem}

Proof of this theorem is based on induction:

\textbf{Base case: } $v_0$, which lies in $c_0$, must lie on the optimal path and is therefore a near-optimal node.

\textbf{Induction step: } Assume that cell $c_i$ contains a near-optimal node $v_i$ on iteration $j$.
From Theorems \ref{thm:select} - \ref{thm:keep}, the probability on any given iteration of propagating $v_i$ into $c_{i+1}$ resulting in a near-optimal node is $P_s P_p > 0$.
Therefore, the probability that at any later iteration $k > j$ a near optimal node exists in $c_{i+1}$ is at least $1-(1-P_sP_p)^{k-j}$. 
As the number of iterations goes to infinity, 
\begin{equation}
\lim_{k \rightarrow \infty}{\big(1-(1-P_sP_p)^{k-j} \big)} = 1.
\end{equation}
Therefore a near-optimal node will exist in $c_{i+1}$ asymptotically almost surely given a near-optimal node exists in $c_i$.
Applying this logic to the finite covering cell sequence, as the planner iterates it will populate each cell in the sequence with a near-optimal node.
Once the algorithm generates a near-optimal node inside the last cell of the sequence that is also inside the goal region, it has generated a near-optimal solution trajectory.
The RDG algorithm will therefore generate a near-optimal solution trajectory asymptotically almost surely.
\qed

\section{Exploration Bias}
\label{sec:Ebias}
One important property of a planner is its ability to generate exploration bias.
The purpose of this section is to examine the methods that the RDG algorithm uses to generate exploration bias and compare them against other commonly used methods. 
The RDG planner generates exploration bias in two ways: biased node selection and cost-based node evaluation. 
Biased node selection refers to the ability of the node selection function, Algorithm \ref{alg:nodeselection}, to select nodes that are more likely, when propagated, to result in trajectories that will terminate in unexplored regions.
The cost-based node evaluation method in Algorithm \ref{alg:graphrevision} keeps the set of selectable nodes small and uniformly spread, which prevents dense clusters of nodes from dominating sampling.

\subsection{Ablation Study}
The exploration bias of the sub-algorithms in RDG was evaluated with an ablation study in a simplified planning environment using simplified planning algorithms.
Each test planner was composed of three components: a state selector, a state propagator, and a node evaluation mechanism.
The state selectors tested were random node selection, the weighted random frontier method, and a traditional Voronoi biasing algorithm created with a nearest neighbor search designed to mirror that of SST\cite{Li2016AsymptoticallyPlanning}.
Each planner used an identical state propagator which generated a trajectory for a set distance in a random direction in two dimensions.
The resulting node was then either always kept (without cell evaluation) or evaluated similar to Algorithm \ref{alg:graphrevision} with a cell grid (with cell evaluation).

The resulting algorithms are described as follows:
\begin{LaTeXdescription}
\item[Random Node (A)] {A random node in the graph is selected and the resulting propagation is then added to the graph. This algorithm is referred to as the ``Naive Random Tree" in \cite{Li2016AsymptoticallyPlanning}.}
\item[WRF With Random Cell Node (B)] {A node is selected by first selecting a cell via the WRF method, then selecting a random node within that cell, regardless of cost. That node is then propagated and added to the graph.}
\item[NNS (C)] {A random state in the state space is sampled and the nearest node in the graph to that state is selected. The node is then propagated and added to the graph.}
\item[Random Node With Cell Evaluation (D)] {A random explored cell in the environment is chosen and its representative active node is selected. The node is then propagated but only added to the graph if it ends in an unexplored cell or is of better cost than every other node in its local cell, as in Algorithm \ref{alg:graphrevision}.}
\item[WRF (E)] {A cell is chosen by the WRF method and the active representative node within that cell is selected. After being propagated, the new node is only added to the graph if it is in an unexplored cell or has the best cost in its local cell.}
\item[NNS With Cell Evaluation (F)] {A random state is sampled and the nearest member of the set of active representative nodes is selected. After propagation it is evaluated and only kept if it is the best cost node in its local cell.}
\end{LaTeXdescription}
Notably, the RDG algorithm itself is distinct from the pure WRF method (E) in that the local cell node selection method is used when $x_{rand}$ lies within an explored cell.
In E, the WRF method is called on every iteration. 

The planners were then initialized in an empty square environment with the root node at the center.
Each algorithm was tested twice, once for 6,750 iterations and once for 30 ms of run time.
6750 iterations and 30 ms of run time produce approximately equal node counts for the NNS with cell evaluation test case, the slowest of the six. 

\begin{figure}
\centering
\includegraphics[width=.9\linewidth]{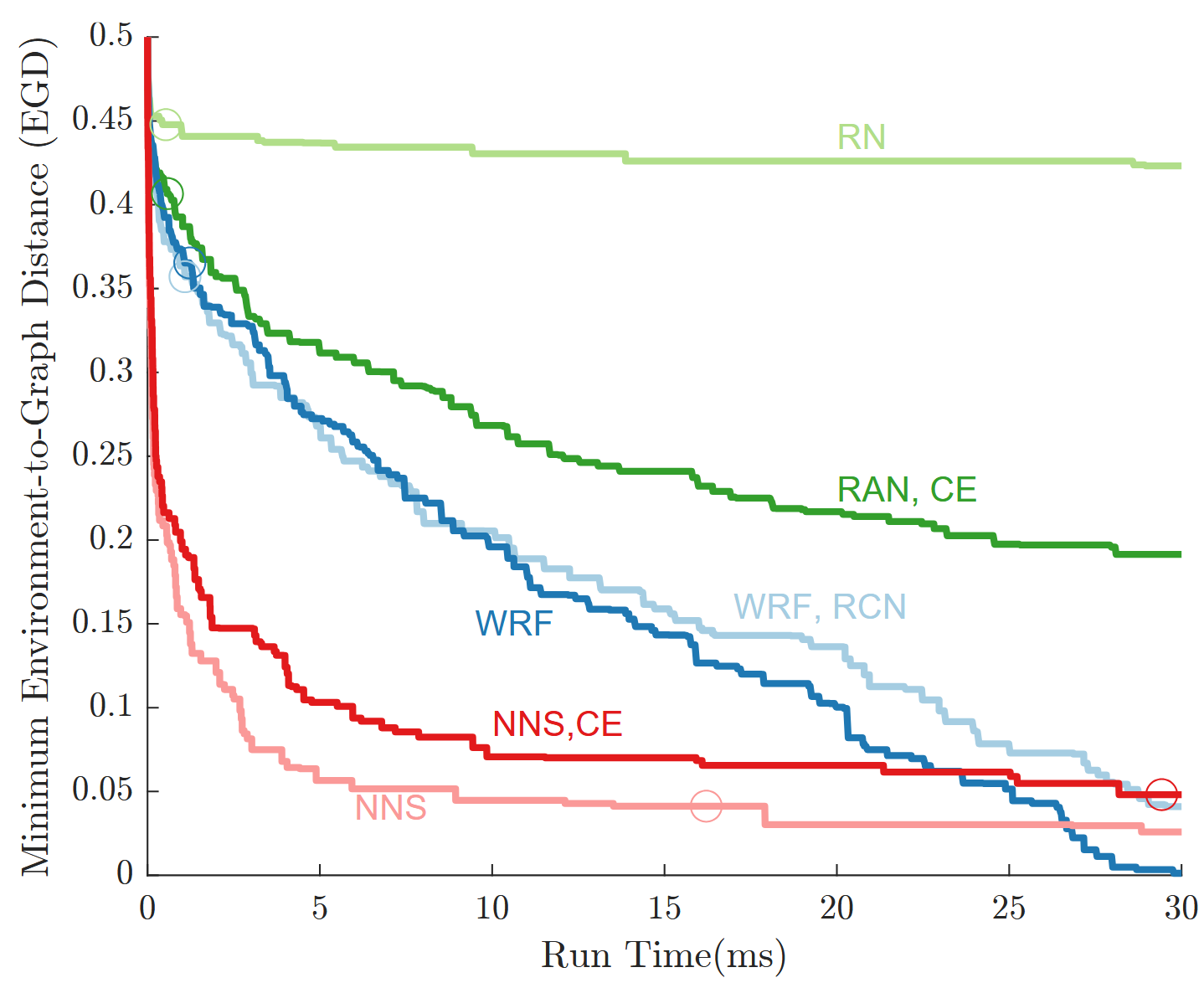}
\caption{Despite exhibiting a weaker exploration bias than the NNS Voronoi bias method, weighted random frontier sampling is fast enough to explore the environment in comparable time scales without invoking an NNS operation. Markers on each curve show the time at which 6750 iterations was reached.}
\label{fig:EGD}
\end{figure}

To quantitatively measure exploration bias, each algorithm run was evaluated by the maximum environment-to-graph distance (EGD) at every iteration.
Maximum EGD measures the greatest distance from any point in the environment to the graph:
\begin{equation}
EGD_{max} = \max_{\forall x \in X, \forall x_i\in G}{\big( ||x-x_i|| \big)}.
\label{eq:EGD}
\end{equation}
Figure \ref{fig:EGD} shows the EGD of each of the six algorithms over the 30 ms run as a percentage of the maximum possible EGD.
Each algorithm begins at approximately 0.5 EGD (the center of the environment) and an EGD of zero corresponds to full environment exploration.

\begin{figure*}
\centering
\includegraphics[width=.85\linewidth]{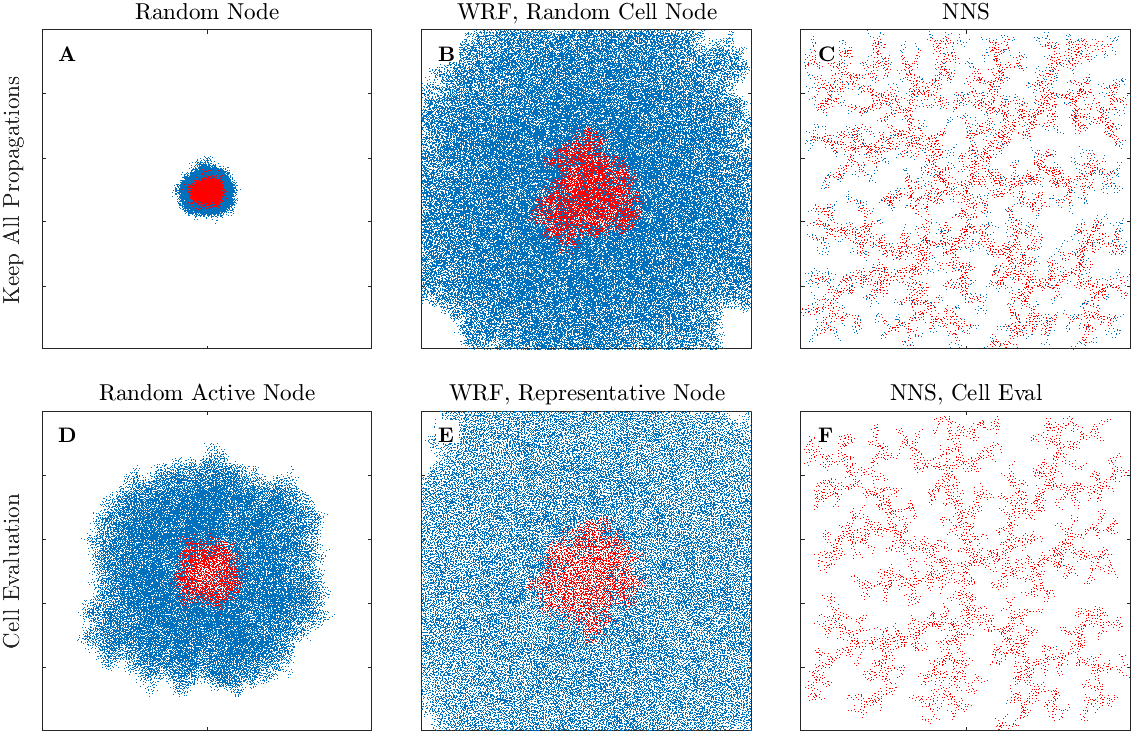}
\caption{Final search trees for all six simplified motion planning algorithms after 6750 iterations (red) and 30 ms of run time (blue) respectively.}
\label{fig:EbiasIterations}
\end{figure*}

\subsection{Exploration Bias Results}
Figure \ref{fig:EbiasIterations} shows graphs generated by each test planner during the 6750 iteration and 30 ms run time tests, respectively.
Each column represents a different node selection method and each row a different graph method.
In the red graphs of Figure \ref{fig:EbiasIterations}, moving from random node selection to the WRF method to an NNS search (left to right) produces increasing levels of exploration bias.
Additionally, adding cell evaluation to random selection and the WRF method yields a slight increase in exploration, seen in frames D and E. 
When run time is fixed instead of number of iterations (blue) the WRF method performs much better than both the random node and NNS methods in terms of saturating the environment.
Like the iteration controlled test cases, the addition of cell evaluation assists both the WRF method and the random node method in exploration.
In both cases, the NNS method did not benefit from the addition of cell evaluation. Qualitatively, these results show that the nearest-neighbor search methods provide better exploration bias when iteration count is fixed while the weighted random frontier methods provide better exploration bias for a fixed amount of time because so many more iterations can occur compared to the NNS methods.

The EGD over time plot of each case, in Figure \ref{fig:EGD}, validates these assessments.
The random node method (RN) shows little decrease in EGD over time.
Selecting a random active node and utilizing cell evaluation (RAN, CE), yields much better exploration.
The two WRF methods (WRF, RCN and WRF) perform similarly in EGD and are able to catch and even surpass the NNS methods close to the 30 ms mark.
The NNS methods (NNS and NNS, CE), while exhibiting very strong exploration bias initially seen as a steep slope on the EGD plot, do not iterate fast enough to achieve complete environmental saturation in the time and iterations given for this problem. 
They do, however, have excellent coverage of the environment for the number of nodes in each graph. 
As graphs and state spaces expand in both size and dimensionality, it is likely that the relative performance of the WRF methods to the NNS methods will increase as well.
Running each iteration in constant time, WRF has better scalability than the NNS-based algorithms, which slow down as node counts increase.
While the addition of cell evaluation to the WRF method does not contribute greatly to exploration bias, it is necessary for asymptotic near-optimality and allows planners to improve solution path cost over time.

\section{Algorithm Evaluation}
The purpose of this section is to evaluate the performance of the RDG algorithm in complex, kinodynamic environments.
RDG is compared to two other algorithms in its class of sparse, forward-propagating, asymptotically near-optimal, kinodynamic motion planners, SST and DIRT \cite{Li2016AsymptoticallyPlanning, Littlefield2018EfficientRegions}.
Because a single-threaded implementation cannot be fairly evaluated against a GPU parallelized one, Kino-PAX was excluded from experimental comparison despite similar algorithmic properties \cite{Perrault2025Kino-PAX:Planner}.

\subsection{Planner Evaluation  Setup}

\begin{figure*}
\centering
\includegraphics[width=.98\linewidth]{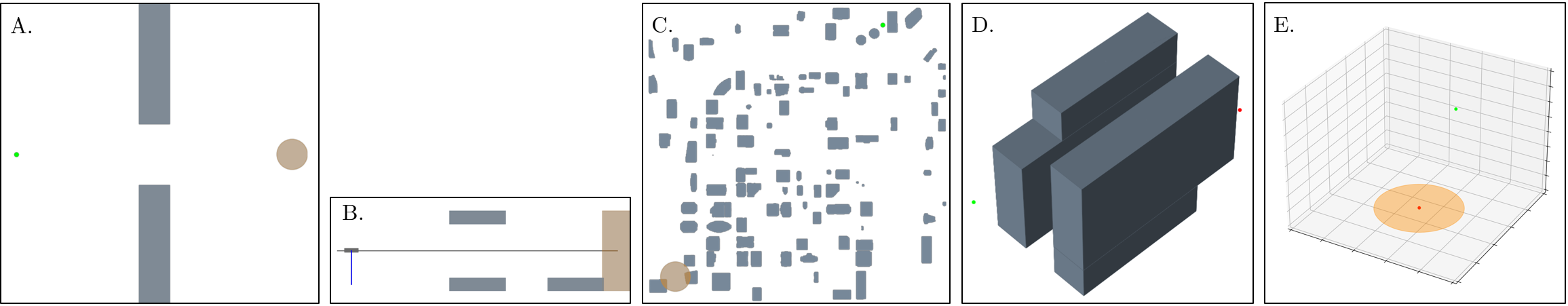}
\caption{(A) 2-D Kinematic Point (B) Fixed wing drone navigating simulated urban environment (C) Cart-mounted underactuated pendulum (D) 6-DOF quadrotor navigating 3-D obstacles (E) 10-DOF rocket landing on a pad.}
\label{fig:envs}
\end{figure*}

Five test cases with different environments and dynamic systems were used.
In each test case, 100 planner instances were run independently for 100 seconds each. 
SST was implemented using the Open Motion Planning Library (OMPL) and the control implementation of SST \cite{Kingston2019ExploringPlanning}.
Despite not being required for all of the problems, the built in OMPL ODE solver was used in all five test cases with the default fourth-order Runge-Kutta integrator. 
RDG also used the same integrator for propagation in all five cases. 
SST was tuned through setting parameters $\delta_{BN}$ and $\delta_s$ so that the ratio of the two radii were similar to ratios used in \cite{Li2016AsymptoticallyPlanning}.
The RDG and SST algorithms were calibrated to each other by setting cell size and witness and best-near radii respectively to equate dynamic clearance of solution paths.
Functionally, controlling for cell size and witness radius by equating dynamic clearance allows both planners to guarantee that they will both be able to find paths with costs that are bounded by the same function $h(\delta,c^*)$.

DIRT was implemented using similar architecture to that of RDG with an identical state propagator feeding the blossom algorithm.
While theoretical properties have not yet been proven on the pruning variant of DIRT, DIRT-prune was used due to its improved performance in comparison to the standard DIRT implementation. 
Admissible heuristics were constructed for each test case consisting of task-space distance from the system to the boundary of the goal region divided by the highest achievable speed for the system in question.
For problems optimizing for travel time, this method guarantees that the heuristic does not overestimate the cost from any given state to the goal region, while still providing strong guidance and not requiring excessive engineering effort to construct. 
DIRT-prune specifically was used for this implementation, as it displayed minor performance improvements on average in the work which proposed the DIRT algorithm \cite{Littlefield2018EfficientRegions}.

\subsection{Test Environments}

Both algorithms were evaluated on one simplified benchmark and four complex kinodynamic motion planning problems. 
Test case environments can be seen in Figure \ref{fig:envs}.
\begin{LaTeXdescription}
\item[Kinematic Point:] {A 2-DOF kinematic point model where velocity is the control input. The state variable is the position of the point. Two obstacles create an environment where the optimal solution path can be easily determined. The dynamic clearance of the optimal solution can also be calculated, allowing analysis of the performance of RDG and SST under conditions where their theoretical properties can be determined.}
\item[Cartpole:] {A 4-DOF cart and pendulum system modeled in \cite{Papadopoulos2014AnalysisSystems}. The free-swinging pendulum is able to be moved by applying horizontal force to the cart body. State variables are cart position and velocity and pendulum angle and angular velocity. The system is tasked with navigating obstacles placed above and below the cart.}
\item[Fixed-Wing Aircraft:] {A 5-DOF fixed wing aircraft dynamics model from \cite{Beard2012SmallPractice}. State variables are position, roll and yaw angles, and airspeed. The aircraft navigates a two-dimensional map of downtown Chicago at a fixed altitude of approximately 100 m. A wind field flowing North was generated through the environment using the CFD software Flowsquare, a free, integrated two-dimensional computational fluid dynamics program \cite{YukiMinamoto2013FlowsquareSoftware}.}
\item[Quadrotor:] {A 6-DOF 3-D quadrotor model with acceleration as the system input. State variables are position and velocity. The system is tasked with traversing two small gaps in a pair of three dimensional obstacles. }
\item[Rocket:] {A 10-DOF 3-D model of a rocket controlled by a thrust input and two engine gimbal angles. State variables are position, pitch, roll, velocity, pitch rate, and roll rate. The system is tasked with landing the rocket on a landing pad with low velocity from an initial state with both position and velocity errors.}
\end{LaTeXdescription}

\subsection{Results}

\begin{figure*}
\centering
\includegraphics[width=0.99\textwidth]{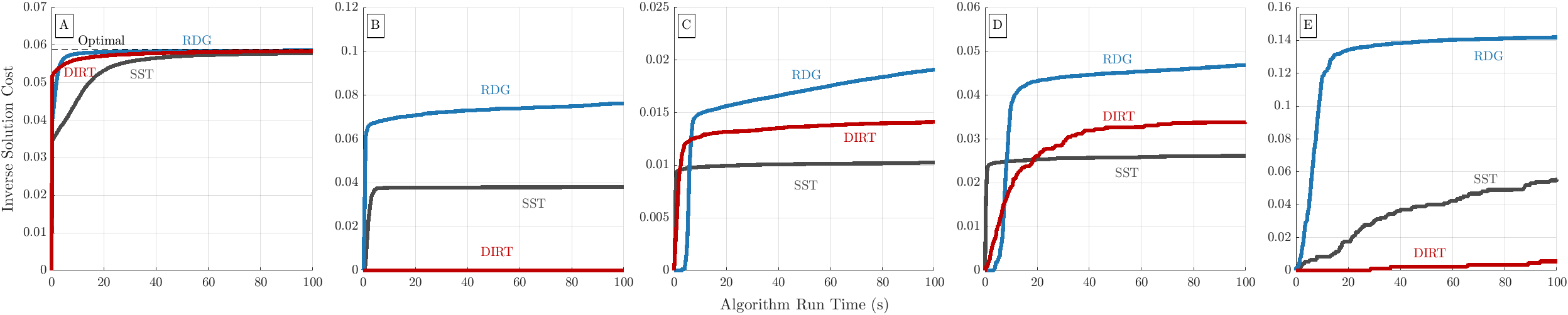} 
\caption{Solution quality of the RDG, SST, and DIRT algorithms over 100 independent 100 second runs. Averages are shown as bold curves. (A) Kinematic Point (B) Cartpole (C) Fixed-Wing (D) Quadrotor (E) Rocket}
\label{fig:resultstime}
\end{figure*}

Figure \ref{fig:resultstime} shows the solution quality, defined as the inverse of solution cost, averaged over all runs vs planning time, where the solution quality of a run at time $t$ is the highest quality among all solution trajectories found by the algorithm at or before time $t$.
Solution quality is used so that outlier-sensitive statistics such as the mean remain useful even when solutions have not yet been found for certain planner instances and their best solution cost is stored as infinite.
The resulting curves are the primary metric of interest to the experimental evaluation.
The five test cases are shown in increasing DOF order in the subplots of Figure \ref{fig:resultstime}.
In each, the RDG algorithm is shown in blue, SST in gray, and DIRT in red. 
In Figure \ref{fig:resultstime}.A, the quality of the optimal solution trajectory is shown as a dashed line. 

\begin{figure*}
\centering
\includegraphics[width=0.99\textwidth]{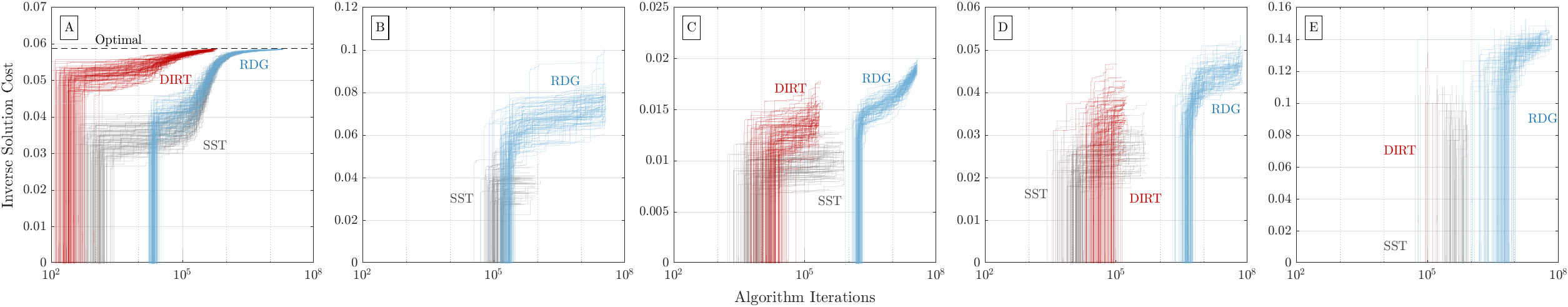}
\caption{Solution quality measured by algorithm iteration. (A) Kinematic Point (B) Cartpole (C) Fixed-Wing (D) Quadrotor (E) Rocket}
\label{fig:resultsiter}
\end{figure*}
Solution quality per iteration is shown in Figure \ref{fig:resultsiter} on a semi-log plot.
Due to differences in final iteration counts, individual runs are shown as individual traces, measured at points where new solutions are found.
The mean of such runs is not included due to differences in algorithm termination iteration that are not present in a plot measured over run time. 

\begin{figure*}
\centering
\includegraphics[width=0.99\textwidth]{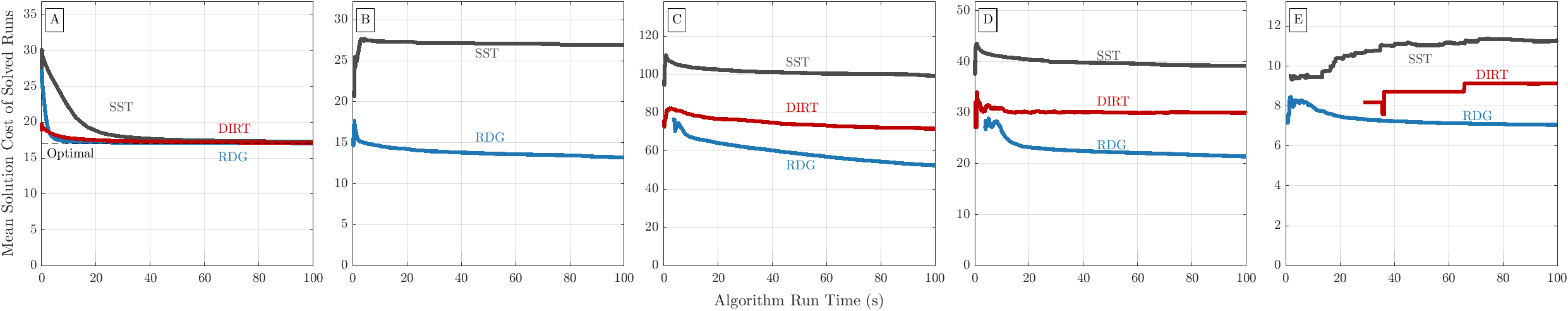}
\caption{Solution cost values averaged over all runs that found solutions. 
(A) Kinematic Point (B) Cartpole (C) Fixed-Wing (D) Quadrotor (E) Rocket}
\label{fig:solvedcost}
\end{figure*}
Mean solution cost averaged over solved runs only is shown in Figure \ref{fig:solvedcost}.
In these plots, cost is not necessarily monotonic, as high-cost first solutions can potentially raise the mean cost over all runs. 
Similar to Figure \ref{fig:resultstime}, in panel A, the optimal solution cost is shown as a dashed line. 

\begin{figure*}
\centering
\includegraphics[width=0.99\textwidth]{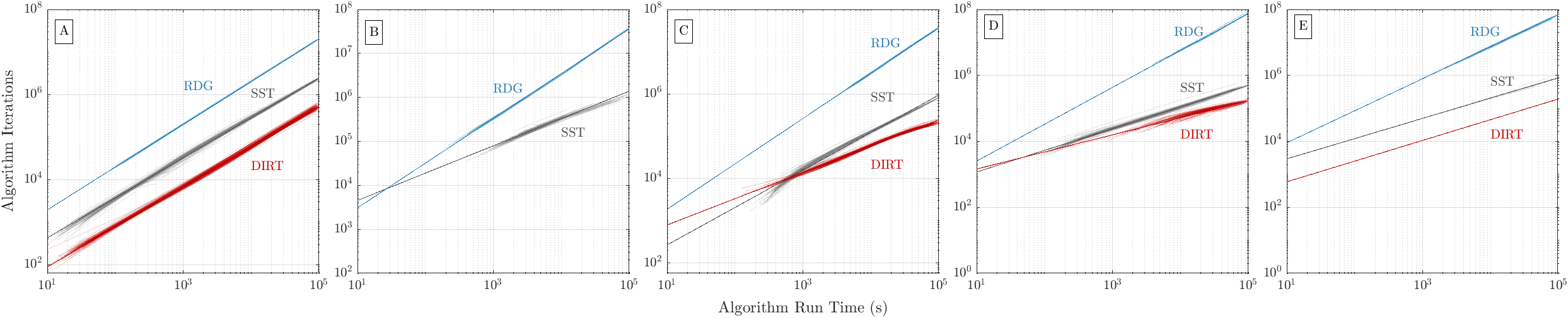}
\caption{Algorithm iterations over time recorded when new solutions are found. Least-squares fits of the log-log data are shown for each algorithm. 
(A) Kinematic Point (B) Cartpole (C) Fixed-Wing (D) Quadrotor (E) Rocket}
\label{fig:iterrate}
\end{figure*}
Finally, planner iteration is plotted over run time for each of the planner runs on log-log axes in Figure \ref{fig:iterrate}. 
Each curve is recorded over the range of run times and iterations where new solutions are found.
Each set of data was fit to a power-law curve, with the power-law exponents also given in Table \ref{table:iterfit}. 
An algorithm with exactly constant iteration run time would have a power law with an exponent of 1. 
An exponent greater than 1 indicates an exponential increase in speed while an exponent less than 1 indicates logarithmic slowdown.

\begin{table}
\centering
\resizebox{\linewidth}{!}{
\begin{tabular}{| c || c | c | c | c | c |} 
 \hline
ALG & A & B & C & D & E\\
\hline
RDG & 1.0052 & 1.0173 & 1.0753 & 1.1176 & 0.9714 \\
\hline
SST & 0.9411 & 0.62149 & 0.88734 & 0.65723 & 0.6141 \\
\hline
DIRT & 0.9443 & N/A & 0.6227 & 0.5192 & 0.6286 \\
\hline
\end{tabular}}
\caption{Power-law exponents for algorithm iteration rate curve fits.}
\label{table:iterfit}
\end{table}

Descriptive statistics are displayed in Table \ref{table:results}. 
Success rate is determined by the number of runs that found feasible solutions within the allotted 100 seconds of planning time. 
Median $t_0$ is the median time before a first solution was found.
The median was used as a descriptive statistic as $t_0$ was found to be right skewed by outlier runs that were late to finding initial solutions.
Solution cost change measures the change in solution cost for each run as a percentage of the initial solution cost. 
The median of solution cost changes across all runs that found a solution is recorded in the table.
Similarly, the median was used due to the skew of the distribution of values.
Mean final cost is the average of the final solution cost of all runs that found a solution.
Final cost standard deviation (SD) is the standard deviation of all such final solution cost values. 
Cliff's $\delta$ is a measure of effect size between two distributions that ranges from $[-1,1]$, with zero meaning stochastic equality and $1$ or $-1$ meaning that one distribution dominates the other \cite{Cliff1993DominanceQuestions}.
In this case, a value of 1 for an algorithm indicates that every run for that algorithm finished with a cost less than that of every RDG run and $-1$ indicates that every RDG run had a lower cost than every run of the algorithm in question.
For each measure, the best algorithm for that test case is highlighted.

\begin{table*}
\centering
\resizebox{\textwidth}{!}{
\begin{tabular} {|c c||C{7.2em}|C{7.2em}|C{7.2em}|C{7.2em}|C{7.2em}|C{7.2em}|}
\hline
Test Case & Algorithm & Success Rate & Median $t_0$ (ms) & Median Solution Cost Change (\%) & Mean Final Cost (s) & Final Cost SD (s) & Cliff's $\delta$ vs RDG\\
\hline

\multirow{3}{4em}{Kinematic Point} 
& RDG & \cellcolor{lightgray} 100/100 & 110 & 36.55 & \cellcolor{lightgray} 17.08 &  \cellcolor{lightgray} 0.0169 & \cellcolor{lightgray} 0  \\
& SST & \cellcolor{lightgray} 100/100 & 31 & \cellcolor{lightgray} 43.28 & 17.29 & 0.0446 & -1.00 \\
& DIRT & \cellcolor{lightgray} 100/100 & \cellcolor{lightgray} 28 & 13.21 & 17.13 & 0.0541 & -0.56 \\
\hline
\multirow{3}{4em}{Cartpole} 
& RDG & \cellcolor{lightgray} 100/100 & \cellcolor{lightgray} 587 & \cellcolor{lightgray} 27.27 & \cellcolor{lightgray} 13.22 & \cellcolor{lightgray} 1.0849 & \cellcolor{lightgray} 0 \\
& SST & \cellcolor{lightgray} 100/100 & 1862 & 2.48 & 26.95 & 4.4512 & -1.00 \\
& DIRT & 0/100 & N/A & N/A & N/A & N/A & N/A\\
\hline
\multirow{3}{4em}{Fixed-Wing} 
& RDG & \cellcolor{lightgray} 100/100 & 5599 & \cellcolor{lightgray} 29.4 & \cellcolor{lightgray} 52.42 & \cellcolor{lightgray} 0.8986 & \cellcolor{lightgray} 0 \\
& SST & \cellcolor{lightgray} 100/100 & \cellcolor{lightgray} 430 & 7.89 & 99.34 & 13.9667 & -1.00 \\
& DIRT & \cellcolor{lightgray} 100/100 & 1346 & 11.35 & 71.58 & 7.4331 & -1.00 \\
\hline
\multirow{3}{4em}{Quadrotor} 
& RDG & \cellcolor{lightgray} 100/100 & 7589 & \cellcolor{lightgray} 25.91 & \cellcolor{lightgray} 21.41 & \cellcolor{lightgray} 1.1045 & \cellcolor{lightgray} 0 \\
& SST & \cellcolor{lightgray} 100/100 & \cellcolor{lightgray} 295 & 11.22 & 39.20 & 6.0823 & -1.00 \\
& DIRT & 99/100 & 8278 & 7.45 & 29.93 & 4.6456 & -0.97 \\
\hline
\multirow{3}{4em}{Rocket} 
& RDG & \cellcolor{lightgray} 100/100 & \cellcolor{lightgray} 6176 & \cellcolor{lightgray} 14.64 & \cellcolor{lightgray} 7.05 & \cellcolor{lightgray} 0.1810 & \cellcolor{lightgray} 0 \\
& SST & 61/100 & 29533 & 0 & 11.30 & 2.0862 & -1.00\\
& DIRT & 5/100 & 65824 & 0 & 9.13 & 0.9454 & -1.00\\
\hline

\end{tabular}
}
\caption{Evaluation results for all planner runs. Best algorithm per column is highlighted.}
\label{table:results}
\end{table*}

\subsection{Discussion}
From the inverse solution cost plots in \ref{fig:resultstime}, RDG can be seen to outperform both SST and DIRT on average in every test case. 
The explanation for the gap in performance can be found with further analysis of Figures \ref{fig:resultsiter}-\ref{fig:solvedcost} and Table \ref{table:results}.

RDG's advantage is driven primarily by its substantially higher iteration rate.
Because the node selection and graph revision methods described in Section III operate in constant time with respect to the size of the tree, RDG sustains a much higher rate of iterations per unit of wall-clock time than either SST or DIRT.
The NNS and FRNNS operations that SST and DIRT require in their node selection and graph revision steps actively slow the algorithms significantly in all but the simplest test case. 
This slowdown can be seen visually in Figure \ref{fig:iterrate} and quantified in Table \ref{table:iterfit}.
The power-law exponents fit to both the SST and DIRT data are both significantly less than one on the kinodynamic test cases, while RDG maintains a rate that is in most cases modestly superlinear.

The reason that RDG maintains a power-law exponent over one in the first four test cases is because of the switch from exploration to refinement that occurs naturally.
As RDG's tree expands and covers more of the state space, the WRF method is called less and less frequently.
While both the WRF method and its refinement alternative are in $O(1)$ time complexity, querying the WRF takes significantly longer than the lookup associated with selecting an explored cell. 
Therefore, in the first four test cases, as the cell grid saturates, the WRF is queried less frequently and the iteration rate of the algorithm tends to accelerate. 
RDG's iteration rate exponent falls slightly below one only in the Rocket test case.
While RDG's performance relative to that of SST and DIRT improves with dimensionality, RDG's own iteration rate shows the first signs of degradation at 10-DOF, plausibly due to hash-map-based cell storage, a consequence of grid sparsity at higher dimensionality.
Behavior beyond 10 DOF was not evaluated in this work.

A clear consequence of the sampling-rate disparity is the low success rate of SST and especially DIRT on the Rocket test case (Table \ref{table:results}).
While RDG is able to find a solution on every run, SST only has a 61\% success rate and DIRT only has 5\%. 
For SST, the lower success rate is coincident with its lowest power-law exponent.
Its second lowest power law exponent also happens to correspond to the test case where SST was able to improve the second least on its solutions: the Cartpole case.
For DIRT, however, the low success rate on the Rocket case is distinct from the total lack of solutions found on the Cartpole case, where a DIRT-specific mechanism led to degenerate trajectory sampling.
This behavior is covered in detail in Section VI.\textit{E}.

The iteration rate advantage of RDG is large enough to overcome the disadvantage in per-iteration efficiency.
As seen in Figure \ref{fig:resultsiter}, RDG usually requires several orders of magnitude more iterations to reach solutions with costs comparable to those found by SST and DIRT.
This result is consistent with the weaker exploration bias per iteration of the WRF method relative to FRNNS- and NNS-based node selection methods in SST and DIRT respectively, a result which is corroborated by the median solution times in Table \ref{table:results}, most of which do not favor RDG.
RDG's overall solution quality advantage is therefore not created by higher-quality iterations, but by sufficient iteration volume to compensate for the relative iteration inefficiency.

The iteration volume advantage also translates into improved trajectory tree refinement, both before and after initial solutions are found. 
Figure \ref{fig:solvedcost} restricts the averaged cost to only runs that have found a solution.
RDG not only finds initial solutions that are of lower cost than those of SST and DIRT, but also improves those solutions more than either SST or DIRT.
Inspecting the median solution cost change in Table \ref{table:results} shows that the typical RDG run improves upon its solution more than double that of the typical SST or DIRT run in all test cases but the Kinematic Point (excluding DIRT Cartpole where no solutions and therefore no improvements were present).
The reason for this one outlier is that median solution cost change is also affected by initial solution quality.
While SST is able to improve its solutions more on a percentage basis than RDG in this individual test case, it ends up with an average solution cost that is over three times further from the optimal solution cost for the problem (0.29 vs 0.08 difference for an optimal cost of 17).
In other test cases, SST and DIRT perform little to no refinement of their trees after initial solutions are found, in part because their trees have so many nodes that the iteration rate of the algorithms has slowed too much for sufficient trajectory sampling volume.

In fact, in many SST and DIRT runs, the last solution found is at a lower iteration count than the number of iterations required for RDG to find its first solution.
In Figure \ref{fig:resultsiter}, this manifests itself as physical separation between the SST/DIRT and RDG curves on the x-axis. 
When two planners utilize the same state propagator, the ability to sample high quality paths is in some part dependent on the number of random samples that are generated.
In this metric, RDG excels and is able to utilize its sampling volume advantage to form high quality solution paths that SST and DIRT simply did not sample enough paths to match. 
The true difference in final cost distribution can be seen via the Cliff's $\delta$ measure in Table \ref{table:results}.
In multiple test cases, SST and DIRT have Cliff's $\delta$ values of exactly -1, meaning that the solution cost of every single RDG run was lower than the solution cost of every single SST/DIRT run.
Even in the cases where Cliff's $\delta$ was not exactly -1, -0.56 and -0.97 are both values that signify strong effects.
Even though the final cost values in the Kinematic Point test case are very close to each other, the Cliff's $\delta$ values for SST and DIRT show the difference in distributions.

In addition to mean solution quality, in which RDG performed best on every test case, RDG also tends to produce paths with much narrower cost distributions, seen in the final cost SD measure of Table \ref{table:results}.
RDG uses a grid discretization that is of predetermined size at the start of a problem, meaning that all runs of the same test case use the same grid discretization.
Identical grid discretizations may reduce potential variability in terminal solution quality.

When compared directly to SST, RDG's improved mean solution quality can be attributed almost directly to its iteration rate advantage.
In fact, the power-law exponent of SST correlates directly with its median relative solution improvement (Spearman $\rho = 0.90$). 
While the small sample size of five test cases means that this figure should not be treated as a formally tested hypothesis, it corroborates the conclusions drawn from Figures \ref{fig:resultsiter}-\ref{fig:iterrate}.
Comparing to DIRT, a similar relationship exists between the slowdown and the gap in solution quality between DIRT and RDG, where the difference is exactly monotonic with DIRT's power-law exponent, excluding the Cartpole test case.
Again, the small sample size precludes a formal conclusion from this correlation alone. 

The zero success rate of the DIRT algorithm on the Cartpole test case is an unexpected finding given the relative success of the algorithm in the work in which it was presented, where it outperformed SST in every test case \cite{Littlefield2018EfficientRegions}.
This failure is attributed to the inability of the heuristic used in this problem, distance to goal divided by maximum cart speed, to capture the energy of the system, a quanity important for sampling swing-up maneuvers. 
Additionally, the design choice to have DIRT create DIRs in task space rather than state space compromises the ability for the DIRT algorithm to effectively use its state-selection mechanism to explore the state space rather than just the task space.
The failure was found to persist under both the pruning and non-pruning variants, multiple different admissible and stronger but inadmissible heuristics, and varying blossom numbers.

\section{Conclusions}
The RDG algorithm is an asymptotically near-optimal kinodynamic planning algorithm that builds a tree of feasible trajectories using a grid of cells in the state space to allow for asymptotic improvement in solution path quality.
Using grid-based node selection and graph revision sub-algorithms, RDG is able to perform each iteration in constant time complexity, allowing for rapid exploration of the environment and fast convergence towards near-optimal solution trajectories.
The RDG algorithm is shown to be $\delta$-robustly asymptotically near-optimal through an induction proof, which allows it to improve solution quality over time and provides guarantees on path quality.
A study on exploration bias on different components of RDG and other motion planners was performed to provide contrast as to the benefits of using grid-based methods relative to commonly used node selection methods.
RDG was evaluated against the SST and DIRT algorithms, members of the class of asymptotically near-optimal, forward-propagating, kinodynamic motion planners in several test cases, exhibiting up to 104\% and 40\% higher solution quality respectively.

\bibliographystyle{IEEEtran}
\bibliography{references}

\end{document}